\documentclass{article}

\usepackage{arxiv}

\usepackage[utf8]{inputenc} % allow utf-8 input
\usepackage[T1]{fontenc}    % use 8-bit T1 fonts
\usepackage{hyperref}       % hyperlinks
\usepackage{url}            % simple URL typesetting
\usepackage{booktabs}       % professional-quality tables
\usepackage{amsfonts}       % blackboard math symbols
\usepackage{nicefrac}       % compact symbols for 1/2, etc.
\usepackage{microtype}      % microtypography
\usepackage{lipsum}
\usepackage{graphicx}

\title{Low-Resolution Face Recognition In Resource-Constrained Environments\thanks{This paper is under consideration at Pattern Recognition Letters}}

\author{
  Mozhdeh Rouhsedagaht\\
  University of Southern California\\
  Los Angeles, California, USA\\
  \texttt{rouhseda@usc.edu} \\
   \And
  Yifan Wang \\
  University of Southern California\\
  Los Angeles, California, USA\\
  \texttt{wang608@usc.edu} \\
  \AND
  Shuowen Hu \\
  Army Research Lab \\
  Adelphi, Maryland, USA \\
  \texttt{shuowen.hu.civ@mail.mil} \\
  \And
  Suya You\\
  Army Research Lab \\
  Adelphi, Maryland, USA \\
  suya.you.civ@mail.mil\\
  \And
  C.-C. Jay Kuo\\
  University of Southern California \\
  Los Angeles, California, USA\\
  \texttt{jckuo@usc.edu} \\
}

\begin{document}
\maketitle

\begin{abstract}

A non-parametric low-resolution face recognition model for
resource-constrained environments with limited networking and computing
is proposed in this work. Such environments often demand a small model
capable of being effectively trained on a small number of labeled data
samples, with low training complexity, and low-resolution input
images. To address these challenges, we adopt an emerging explainable
machine learning methodology called successive subspace learning (SSL).
SSL offers an explainable non-parametric model that flexibly trades the
model size for verification performance. Its training complexity is
significantly lower since its model is trained in a one-pass feedforward
manner without backpropagation. Furthermore, active learning can be
conveniently incorporated to reduce the labeling cost. The effectiveness
of the proposed model is demonstrated by experiments on the LFW and the
CMU Multi-PIE datasets. 

\end{abstract}

%\linenumbers


\section{Introduction}\label{sec:introduction}

Cloud-based face recognition has reached maturity in recent years. Deep
neural network (DNN) models consisting of millions of model parameters
have been developed and made significant progress. The high performance
mainly relies on several factors: higher input image resolutions, using
an extremely large number of training images, and abundant
computational/memory resources. Take the
models which have achieved high accuracy on the LFW benchmark as an example. DeepFace~\cite{taigman2014deepface}
was trained on a collection of photos form Facebook that contains 4.4M
images. FaceNet~\cite{schroff2015facenet} was trained on the Google
dataset that contains 500M images. SphereFace~\cite{liu2017sphereface}
was trained on the CASIA-WebFace dataset~\cite{yi2014learning} that
contains 0.49M images. Along this direction, many activities are
centered on face images collection and the setup of the required computing
and communication environment. 

We may face an opposite situation in some real-world applications; i.e.,
edge or mobile computing in resource-constrained environments with poor
computing and communication infrastructure, as is often the case in the
field and in operational settings. Such environments demand smaller
model size, fewer labeled images for training, lower training and
inference complexity, and lower input image resolution, partly due to
the need to image individuals at farther standoff distances. Due to
these stringent requirements, DNNs may not be applicable. The goal of
this work is to address these challenges and develop a robust and
transparent non-parametric model that allows graceful performance
tradeoff between resources and performance and is capable of being
easily integrated with active learning to minimize the training sample
number while achieving relatively high accuracy. We adopt an emerging
machine learning system called PixelHop++~\cite{chen2020pixelhop++} to
achieve these objectives.  PixelHop++ is designed based on the
successive subspace learning (SSL) principle and has several unique
characteristics that fit our objectives well. 
\vspace{-1ex}
\begin{enumerate}
\setlength\itemsep{0em}
\item PixelHop++ is a light-weight non-parametric model whose size can
be flexibly adjusted for graceful performance tradeoff. Its training is
conducted in a feedforward one-pass manner and the training complexity
is significantly lower than DNNs. 
\item SSL adopts a statistics-centric principle. It exploits
pixel-to-pixel correlations for dimension reduction to derive image
features. It also analyzes statistics between features and labels to
identify discriminant features. SSL is mathematically transparent.
\item We will incorporate active learning in SSL to select the most
``informative'' samples of a dataset for labeling and reduce the
labeling cost. PixelHop++ is a lightweight model, and it can easily
be integrated with active learning. 
\end{enumerate}
The main contribution of our work lie in the assembly of two effective
tools to address the challenge of face recognition in
resource-constrained environments. Both PixelHop++ and active learning
are existing tools. Yet, to the best of our knowledge, this is the first
time that they are jointly applied to a face biometric problem. We will
demonstrate the power of the integrated solution in the context of face
recognition with extensive experiments. 

The rest of the paper is organized as follows. Related prior work is
reviewed in Sec. \ref{sec:review}. The proposed face recognition
method and its integration with active learning are presented in Sec.
\ref{sec:framework} and Sec. \ref{sec:active}, respectively.
Experimental results are provided in Sec. \ref{sec:experiments}.
Finally, concluding remarks and future work are given in Sec.
\ref{sec:conclusion}. 
 
\section{Related Work}\label{sec:review}

{\bf Face Recognition.} Face recognition has made significant progress
in recent years. Most successful face recognition models use deep
learning technique \cite{taigman2014deepface, sun2014deep,
schroff2015facenet, parkhi2015deep,liu2017sphereface} and offer high
accuracy on the benchmarking datasets. Although these models are
powerful for high-resolution face recognition, they usually contain tens
of millions of model parameters and require a large amount of training data and
computation resources. A light-weight CNN model called Light CNN was
proposed in \cite{Wu_2018} which is significantly smaller than regular
CNN networks but still has millions of model parameters. 

{\bf Low-Resolution Face Recognition.} In comparison with
high-resolution face recognition, less attention has been drawn to
low-resolution face recognition. Generally, there are two different
settings for this problem: high-resolution to low-resolution, and
low-resolution to low-resolution. In the first setting, the low-resolution probe images are
compared against high-resolution gallery images \cite{biswas2011multidimensional,mudunuri2015low,lu2018deep} while in the second
setting both
probe and gallery images are low-resolution face images \cite{wang2016studying,cheng2018low, ge2018low}. We evaluate
our model under the second setting and compare it with the model
proposed in \cite{ge2018low}. 

{\bf Successive Subspace Learning (SSL).} The main technique in subspace
learning is to project a high-dimension input space to a low-dimensional
output subspace, which serves as an approximation to the original one.
When these dimension reduction operations are performed sequentially, it
leads to successive subspace learning (SSL).  PixelHop++
\cite{chen2020pixelhop++} is a model developed using the SSL principle.
It is proposed for unsupervised representation learning by applying the
channel-wise (c/w) Saab transform to images.  The Saab transform
\cite{kuo2019interpretable} is a multi-stage variant of principal
component analysis (PCA) conducted on images. In each stage, it applies
PCA to pixel blocks and also uses a bias term to avoid the sign-confusion
problem \cite{kuo2016understanding}.  The performance of the Saab
transform can be further enhanced by removing the spatial correlation
between Saab transform outputs in the current stage so that the Saab
transform can be applied to each output channel separately in
the next stage.  PixelHop++ with the c/w Saab transform is proven to
offer an effective multi-stage representation. For example, FaceHop
\cite{rouhsedaghat2020facehop} is recently proposed for gender
classification on gray-scale face images. It uses PixelHop++ for feature
learning. In this paper, besides incorporating chrominance channels of face
images, we add a new module for effective pair-wise feature extraction
to tackle the face recognition problem. 

{\bf Active Learning.} Active learning is used to select the most
informative unlabeled data for labeling so that almost the same accuracy
can be reached with the smallest amount of labeled data. An active
learning method begins with a small amount of labeled data to train a
machine learning model. The model queries the labels of some unlabeled
data based on a query strategy, and the model is retrained by all
labeled data. This process is repeated until we reach the labeled sample
budget. Common query strategies include the entropy
method~\cite{shannon1948mathematical}, the query by committee
method~\cite{seung1992query} and the core-set
method~\cite{sener2017active}.  Active learning was incorporated in
DNNs by Wang {\em et al.}~\cite{Wang_2017}.  An active annotation and
learning framework is introduced in ~\cite{Ye2016} for the face
recognition task. 

%%%%%%%%%%%%%%%%%%%%%%%%%%%%%%%%%%%%%%%%%%
\begin{figure}[t]%[!h]
\begin{center}
\includegraphics[width=0.75\linewidth]{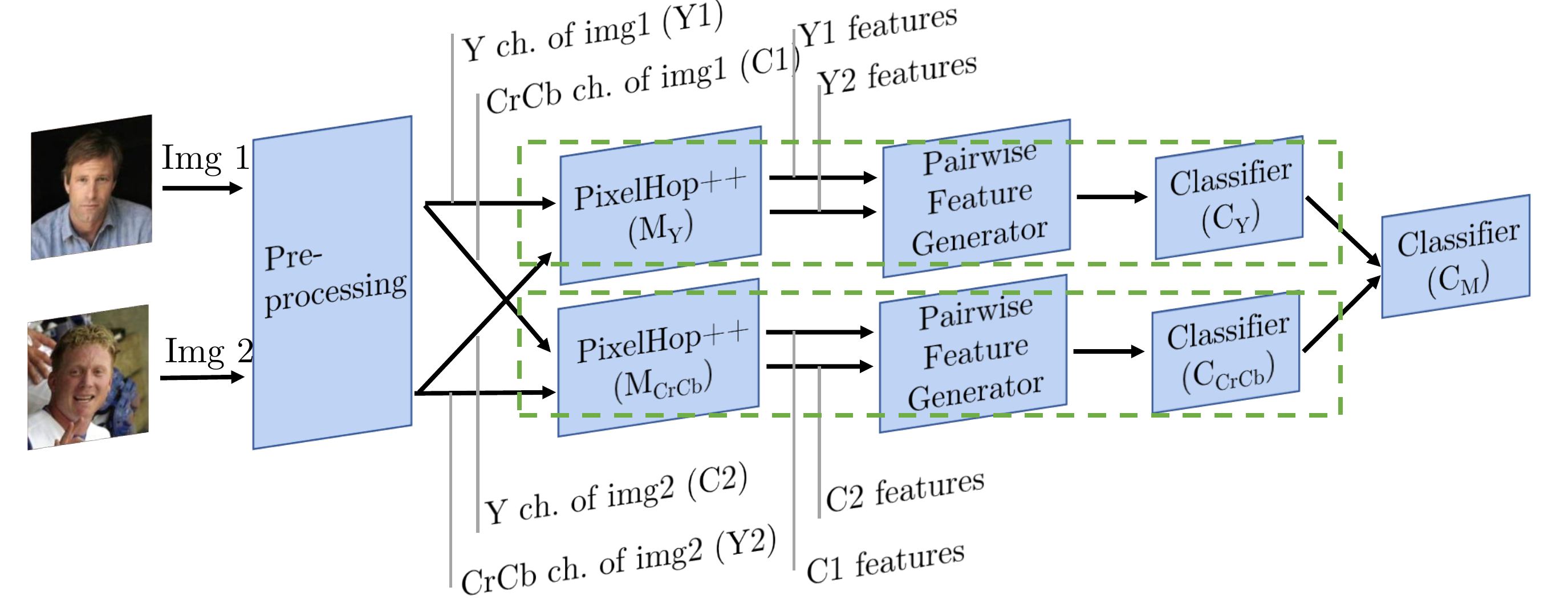}
\end{center}
\caption{The block diagram of the proposed face recognition method.}
\label{fig:blockdiagram}
\end{figure}
%%%%%%%%%%%%%%%%%%%%%%%%%%%%%%%%%%%%%%%%%%

\section{Proposed Face Recognition Method}\label{sec:framework}

The block diagram of the proposed face recognition method is shown in
Fig. \ref{fig:blockdiagram}. As shown in the figure, our model is an
ensemble of two submodels, each of which consists of three components, i.e., PixelHop++, pairwise feature generation, and classifier. We will elaborate the function of each component of the block diagram below. 

\subsection{Preprocessing}\label{subsec:preprocessing}

Several commonly used face processing techniques are adopted in the 
preprocessing block such as
\vspace{-1ex}
\begin{enumerate}
\setlength\itemsep{0em}
\item Applying a 2D face alignment algorithm to the input face images to
reduce the effect of pose variation;
\item Cropping face images properly to eliminate background;
\item Using histogram equalization (HE) to reduce the effect of
different illumination conditions on the Y channel. 
\end{enumerate}
We use the dlib~\cite{dlib09} toolkit for face detection and landmark
localization. Face images are aligned/normalized so that the line
connecting the eye centers is horizontal and all faces are centered and
resized into a constant size of 32×32 pixels. We convert the color space to YCrCb
and, then, feed the Y channel (the luminance component) separately to
one submodel and the Cr and Cb channels (chrominance components) to
another submodel.

\subsection{PixelHop++}\label{subsec:pixelhop++}

%%%%%%%%%%%%%%%%%%%%%%%%%%%%%%%%%%%%%%%%%%
\begin{figure}[t]
\begin{center}
\includegraphics[width=0.7\linewidth]{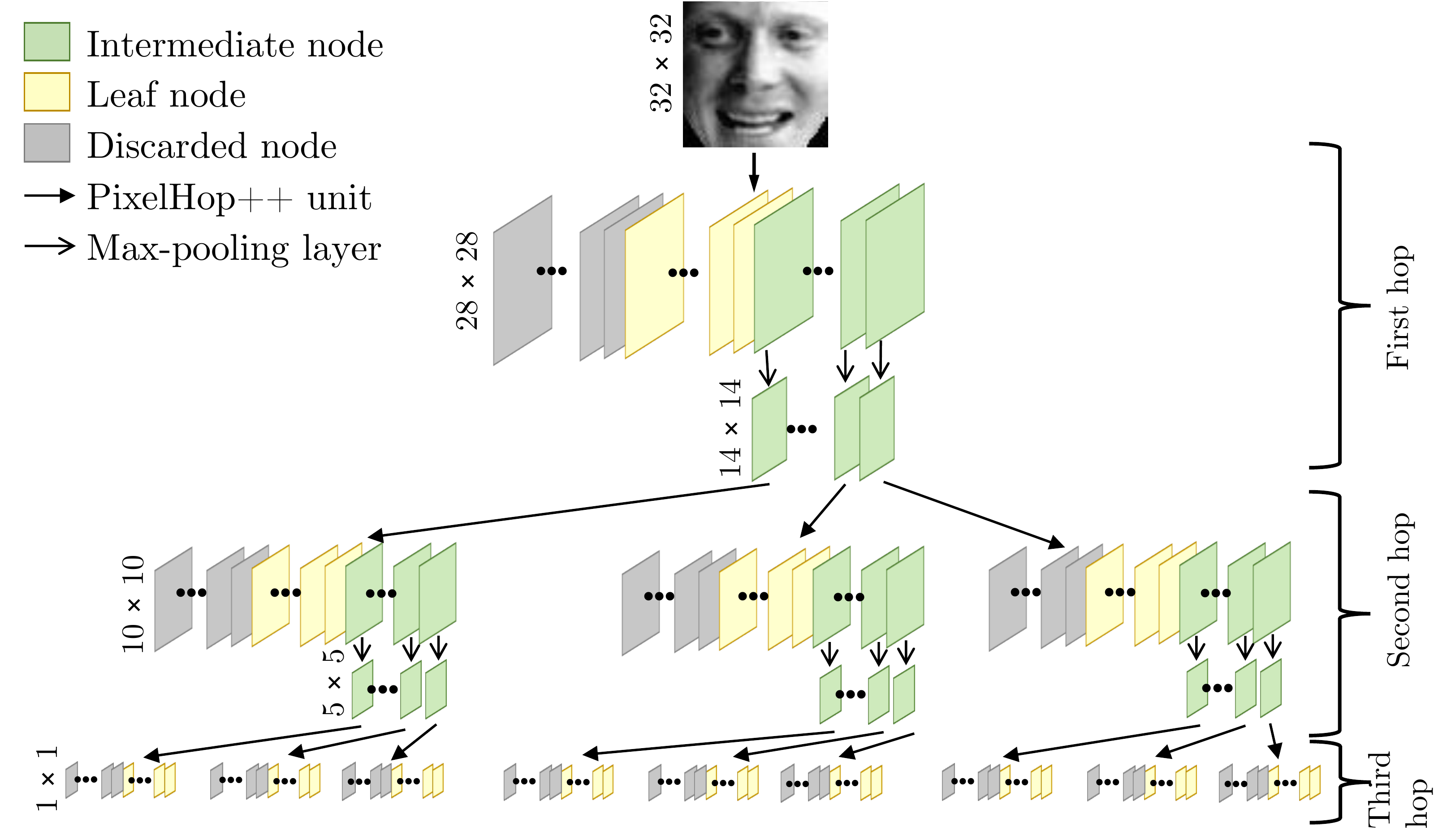}\\
\end{center}
\caption{Illustration of data flow in the three-level c/w Saab transform
in PixelHop++, which provides a sequence of successive subspace
approximations (SSAs) to the input image.} \label{fig:pixelhop}
\end{figure}
%%%%%%%%%%%%%%%%%%%%%%%%%%%%%%%%%%%%%%%%%%

In each submodel, we use a three-level PixelHop++ system, similar to the system shown
in Fig. \ref{fig:pixelhop}. The input to each of the
PixelHop++ systems is a whole face image of resolution 32×32×$K_0$,
where $K_0=1$ and $2$ for the Y channel and CrCb channels, respectively.
Each level of a PixelHop++ system has one PixelHop++ unit followed by (2×2)-to-(1×1) max-pooling. In our model, the PixelHop++ unit of each level operates on blocks of 5×5 pixels with a stride of one.  In the training phase of each PixelHop++ unit, we collect sample blocks from each input channel to derive
Saab kernels for that channel separately. Then, we project each
block on the derived kernels and generate a set of responses for the
central pixel in the block. Since the responses can be positive or
negative, we add a constant bias to all responses to ensure that they
are all non-negative, which explains the name of the ``successive
approximation with adjusted bias (Saab) transform"
\cite{kuo2019interpretable}. 

The first Saab kernel is the unit-length constant-element vector that
computes the local mean of each block which we call the DC component. After
removing the DC, we conduct PCA on residual blocks. Since each block has
25 dimensions, we get one DC component plus 24 AC components, whose
kernels are eigenvectors of the covariance matrix of the collected
blocks, for each channel.%block.
In each level, the corresponding Saab components
of the blocks of each channel form an output channel (or a node in a
tree as shown in Fig. \ref{fig:pixelhop}), e.g. the DC component of blocks extracted from an input channel from one output channel/node.  We divide these nodes into
three groups:
\vspace{-1ex}
\begin{itemize}
\setlength\itemsep{0em}
\item Intermediate nodes: The DC and several leading low-frequency AC
channels will be forwarded to the next level for further energy
compaction. 
\item Leaf nodes: The nodes which are kept at the current level. 
\item Discarded nodes: AC components with very small eigenvalues 
are discarded.
\end{itemize}
As we mentioned, in each PixelHop++ unit we apply channel-wise (c/w) transform to pixel blocks; in other words, for pixel blocks extracted from each individual input channel, an individual Saab transform is applied. The reason we can process channels individually is that all AC channels are orthogonal to the DC channel and all AC responses
are uncorrelated due to PCA. Note that the first PixelHop++ unit of $M_{CrCb}$ is an exception (Cr and Cb channels are not uncorrelated), so we apply Saab transform on blocks of 5×5×2 pixels at this level and obtain 1 DC component and 49 AC components for each pixel block. We should
emphasize that {\em channel separability} is powerful in reducing our
model size and computational complexity. Unlike DNNs, PixelHop++ does
not transform one large 3D (i.e., 2D spatial plus 1D spectral) tensor
but multiple 2D spatial tensors.

Each level of PixelHop++ provides an approximation to the input with
different spatial/frequency tradeoffs. The input is a pure spatial
representation.  The output of Level-3 is a pure frequency
representation. The outputs of Level-1 and Level-2 are hybrid
spatial/frequency representations. A square in Fig.  \ref{fig:pixelhop}
indicates a channel which is the union of all corresponding spatial
locations.  Before max-pooling, the dimensions of the output channels of
Level-1 and Level-2 are 28×28 and 10×10, respectively. Clearly, Level-1
has more spatial detail than Level-2.  The spatial dimension of Level-3
is 1×1. Each intermediate/leaf node at a level indicates a frequency
channel at the corresponding level. 

For channels at each level, we need two hyper-parameters to partition
them into the mentioned three groups.  We use the energy level of a
channel as the criterion. If its energy is less than a cutoff energy
level denoted by $E_C$, the channel is discarded. If its energy is
higher than a forwarded energy level denoted by $E_F$, the channel is
forwarded to the next level.  We normalize the energy of the root node
(i.e., the input image) to 100\%.  The energy level of each
intermediate/leaf node is computed as follows. 
\vspace{-1ex}
\begin{itemize}
\setlength\itemsep{0em}
\item Step 1: Initial DC and AC energy computation for each PixelHop++ unit. \\
Each eigenvalue of the covariance
matrix indicates the energy of its corresponding node. We define the initial energy ($E_{init}$) of each output node as the ratio of its corresponding eigenvalue to the sum of all others eigenvalues related to that PixelHop++ unit. At this step, the sum of the DC and total AC energy values for each PixelHop++ unit is 100\%. 
%This only occurs when we pass the DC channel to the next level. 
\item Step 2: Normalized energy at each node. \\
Based on the first step, we have the energy of an
intermediate/leaf node against its siblings. By traversing the tree from
the root node to a leaf node, the path includes intermediate nodes. The
normalized energy of a leaf node against the root node is the product of
$E_{init}$ values of all visited
nodes (including itself). 
\end{itemize}
Note that by lowering $E_C$ and $E_F$, we can obtain a better
approximation at the cost of higher model complexity. 

\subsection{Pairwise Feature Generation}\label{subsec:feature}

To compare whether two face images are similar or not, we can examine
their similarities at different spatial regions, channels, and
levels. This is feasible because of the rich representations offered by
PixelHop++. % A node at one level corresponds to one frequency channel. 
Note that although the content of an intermediate node will be mainly forwarded to the next level, it does have different spatial/spectral representations at
two adjacent levels. Thus, for
feature extraction, we do not differentiate intermediate/leaf nodes at
each level. 

\begin{figure}[t]%[!h]
\begin{center}
\includegraphics[width=0.45\linewidth]{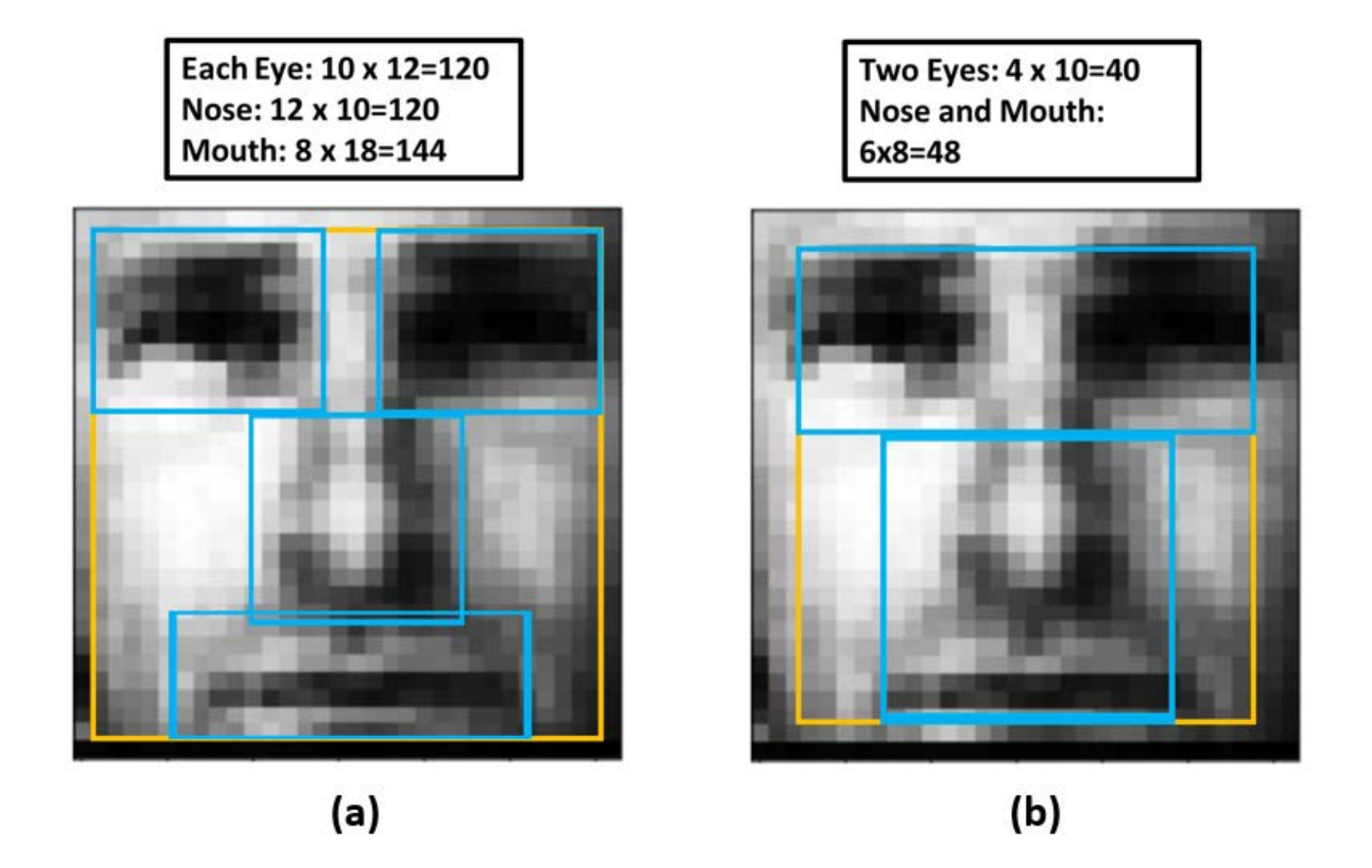}
\end{center}
\caption{Illustration of selected spatial regions of interest (ROIs)
with respect to the input image for frequency channels at (a) Level-1
and (b) Level-2.}\label{fig:ROI}
\end{figure}

There are $K_1$, $K_2$ and $K_3$ nodes at Level-1, Level-2
and Level-3, respectively, as shown in Fig. \ref{fig:pixelhop}. We can
process them individually as detailed below.
\vspace{-1ex}
\begin{itemize}
\setlength\itemsep{0em}
\item {\bf Level-1.} It has the highest spatial resolution 28×28. We
can use it to zoom in on salient regions of the face such as the left eye, the right eye,
the nose, and the mouth of certain dimensions at each channel (see Fig. \ref{fig:ROI}(a)) and flatten them to form vectors. Accordingly, we extract 4 feature vectors from each node at this level (4×$K_1$ feature vectors). 
\item {\bf Level-2.} It has a lower spatial resolution (10×10). We can
still zoom in on unions of salient regions such as one horizontal stripe covering
two eyes and one vertical stripe covering the nose and the mouth at each
channel (see Fig. \ref{fig:ROI}(b)) and form 2 vectors per node accordingly (2×$K_2$ feature vectors). 
\item {\bf Level-3.} It has no spatial resolution. Each leaf node offers
a scalar description of the whole face. %To compute the cosine similarity, we need to group multiple channel outputs into one vector.
In our implementation, we concatenate all $K_3$ nodes into a long
sequence and then group every 10 nodes as one vector. Consequently, we obtain $P$
feature vectors, where $P$ is the floor of $K_3$ divided by 10. 
\end{itemize}

%%%%%%%%%%%%%%%%%%%%%%%%%%%%%%%%%%%%%%%%%%
%%%%%%%%%%%%%%%%%%%%%%%%%%%%%%%%%%%%%%%%%%

To compare the similarity between two face images, we collect corresponding
vector pairs from the same spatial region of the same node
(including intermediate and leaf nodes) at each level and compute two
similarity measures for them - the cosine similarity ($C_k$) and the length ratio
($R_k$) for the $k$-th vector pair. We define the length ratio of two vectors as the ratio of the vector with smaller L2 norm to the vector with larger L2 norm. If two images are similar, their cosine
similarity and length ratio should be close to unity. Otherwise, they
should be farther away from (and less than) one. We observe
experimentally that an individual $R_k$ value is not as discriminant as
$C_k$. Instead, the average length ratio for each spatial region in
Level-1 and -2 as shown in Fig.  \ref{fig:ROI} and the average length ratio for $P$ pairs in Level-3
is more robust and discriminant. 

To summarize, the ultimate feature vector to be fed to the binary
classifier is the concatenation of: 1) 7 average length ratio values (4 extracted from Level-1, 2 from Level-2, and 1 from Level-3),
2) 4×$K_1$ cosine similarity values from four spatial regions and $K_1$
channels, 3) 2×$K_2$ cosine similarity values from two spatial regions
and $K_2$ channels, and 4) $P$ cosine similarity values from groups of channels in Level-3. %that share the same parent node.  
The value of the feature dimension (N) for each submodel
is given in Table. \ref{table:acc1}. 

\subsection{Classifiers}\label{subsec:classifier}

As described in Sec. \ref{subsec:feature}, we extract the feature vector
from the Y channel and the CrCb channels of each pair separately.  For
each pairwise feature, we train a classifier which makes a soft
decision, i.e., the probability for the pair to be match or mismatch.
Then, we feed these two probabilities into a meta classifier for the
final decision.  We use the
LR classifier in our experiments to achieve a smaller model size
although using larger binary classifiers we may achieve higher accuracy.

\section{Integration with Active Learning}\label{sec:active}

The feature generation process in our model is unsupervised, and labels
are only needed for classifier training. As a motivation for active
learning, a scenario of interest is training a model on a mobile agent
when the model has access to unlabeled training samples locally but has
to fetch the label of a limited number of samples from the server
through unreliable low-bandwidth channels. To overcome the communication
constraint, active learning can be used to retrieve labels of most
informative samples. We consider
three active learning methods as explained below. 
\begin{enumerate}
\item \textit{Entropy method}: In each iteration, compute the entropy of each
sample in the pool of unlabeled data $D^u$ and pick the samples with
higher entropy to add to the labeled training data. Higher entropy means
the model is more uncertain about those samples. The entropy for sample
X can be computed using 
\begin{equation} 
entropy(x) = -\sum_{j=1}^{J} p({y_i\mid x})\log p({y_i\mid x}),
\end{equation}
where p($y_i\mid x$) is the probability that sample $x$ belongs to the class
label $y_i$. 
\item \textit{Query By Committee (QBC)}: Instead of training one model,
this method trains several models called a committee. In each iteration,
measure the disagreement between the committee members for each sample
in the pool of unlabeled data and pick the ones with the largest
disagreement values. One of the common disagreement measures is Vote
Entropy which can be computed using
\begin{equation} 
VE(x) = -\sum_{j=1}^{J} \frac{V({y_i})}{C}\log \frac{V({y_i})}{C},
\end{equation}
where $C$ is the number of models, $V(y_i)$ is the number of votes which
label $y_i$ receives from committee members. 
\item \textit{Core-set method}: The core-set selection problem is
choosing $b$ sample points from the pool of unlabeled data that minimize
the maximum distance between each data point remaining in the pool and its
nearest data point in the selected subset. This problem is NP-Hard. A
greedy approach for core-set selection is the k-Center-Greedy algorithm.
In the $i^{th}$ iteration, it selects the samples from $D^u$ with the
maximum distance from their closest sample in the labeled training data
$D^l_{i-1}$. 
\end{enumerate}

\section{Experiments}\label{sec:experiments}

We evaluate the performance of the proposed method by
conducting experiments on two well-known datasets: Labeled Faces in
the Wild (LFW) \cite{huang2008labeled} and CMU Multi-PIE \cite{gross2010multi}. In all the experiments, we use
low-resolution face images of size 32×32 unless otherwise specified. 

The \textbf{\textit{LFW dataset}} is a widely used dataset for face
verification. It consists of 13,233 face images of 5,749 individuals.
For performance benchmarking of different face verification models, it
provides 6,000 face pairs in 10 splits.  We follow the
``Image-Restricted, Label-Free Outside Data" protocol. We choose this protocol as
we use a tool for facial landmark localization in the preprocessing
step. But for training the model, we only use face images in the LFW training set.  A 3D
aligned version of LFW~\cite{ferrari2016effective} is used in
experiments. For data augmentation, we add the pair of horizontally
filliped images of each training pair to the training data. 
 
The \textbf{\textit{CMU Multi-PIE dataset}} contains more than 750,000
images of 337 people recorded in four sessions. For each identity,
images are captured under 15 viewpoints and 19 illumination conditions
with a few different facial expressions. In experiments, we select a
subset of 01 session which contains frontal and slightly non-frontal
face images (camera views 05\_0, 05\_1, and 14\_0) with a neutral
expression and under all existing illumination conditions.

\subsection{Face Verification on LFW}
%%%%%%%%%%%%%%%%%%%%%%%%%%%%%%%%%%%%%%%%%%%%%
\begin{table}[t]
\caption{Comparison of test accuracy of $C_Y$ and $C_CrCb$ and
hyper-parameter settings of $M_Y$ and $M_CrCb$, where $K_1$, $K_2$, and
$K_3$ are numbers of intermediate and leaf nodes at Level-1, Level-2,
and Level-3, $P$ is the number of vectors at Level-3 and
$N=7+4K_1+2K_2+P$ is the feature dimension.} \label{table:acc1}
\begin{center}
\begin{tabular}{cccccccc}
\hline
Input ch.  & $E_C$ & $K_1$ & $K_2$ & $K_3$ & $P$ & $N$ & Acc. \\ \hline
Y & 0.0005 & $18$ & $119$ & $233$ & $23$ & $340$ & 83.47 \\
CrCb & 0.0004 & $19$ & $73$ & $124$ & $12$ & $241$ & 75.89 \\ \hline
\end{tabular}
\end{center}
\end{table}
%%%%%%%%%%%%%%%%%%%%%%%%%%%%%%%%%%%%%%%%%%%%%
{\bf Experiment \#1.} We use the first 90\% of the LFW training pairs
for training the model and the rest of the pairs for testing. The
parameters of each PixelHop++ model and the accuracy of the classifiers
trained on their output features are given in Table. \ref{table:acc1}.  We
set $E_C$ = $E_F$ so that the number of leaf nodes in the first and
second hop is zero. The test accuracy of the meta classifier ($C_M$)
under this setting is 85.33\%. We use the same hyper-parameters in all
experiments. 

%%%%%%%%%%%%%%%%%%%%%%%%%%%%%%%%%%%%%%%%%%
\begin{figure}[t]
\begin{center}
\includegraphics[width=0.45\linewidth]{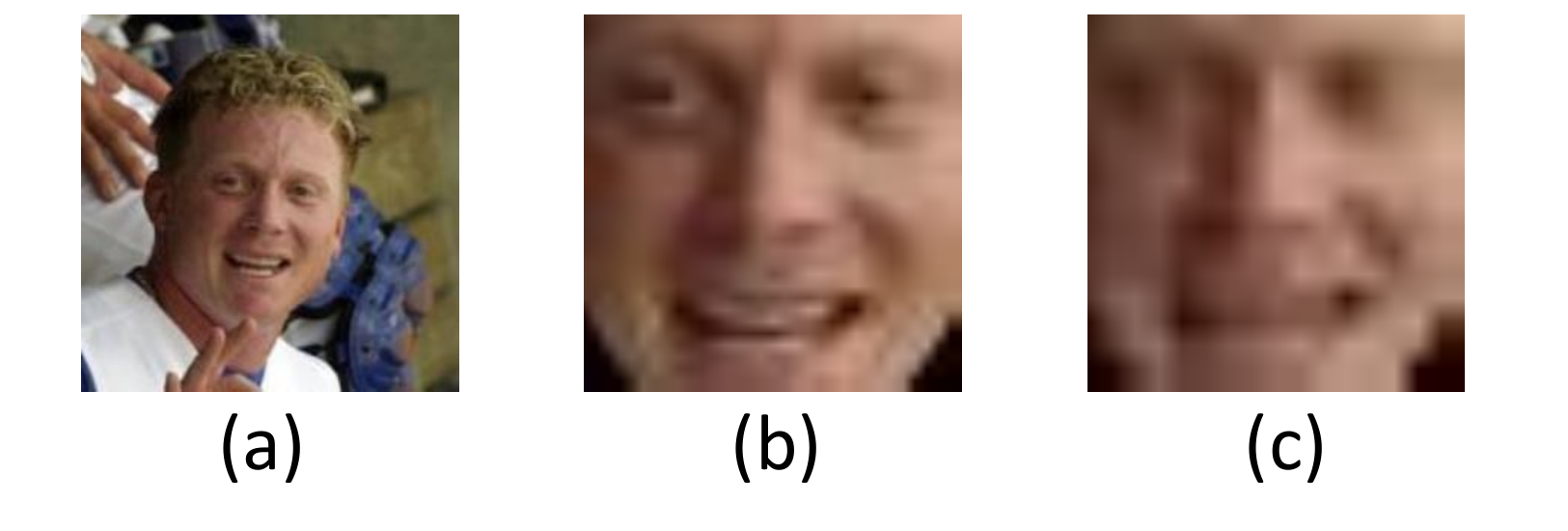}
\end{center}
\caption{Quality of the obtained 32×32 (b) and 16×16 (c) low-resolution
face images compared with the original high-resolution face image (a).}
\label{fig:quality}
\end{figure}
%%%%%%%%%%%%%%%%%%%%%%%%%%%%%%%%%%%%%%%%%%
{\bf Experiment \#2.} To compare our model with the state-of-the-art, we compute the 10-fold cross-validation accuracy for the input
resolution of 32×32 and 16×16. To obtain the 16×16 face images, we
down-sample images to 16×16 and then resize them back to 32×32.  The
quality of the obtained face images is compared in Fig.
\ref{fig:quality}.  To the best of our knowledge, there is no
low-resolution to low-resolution face recognition model which has
reported accuracy under the ``Image-Restricted, Label-Free Outside Data"
protocol on LFW. To this end, we compare our model with the SKD
(Selective Knowledge Distillation) model \cite{ge2018low} which is a
recent state-of-the-art model for low-resolution face recognition using
unlimited training data. 

We have compared our model, SKD without distillation, and SKD in terms of accuracy, the number of parameters, and the
number of training images for each input resolution in Table.
\ref{table:lfw}. We see that for 16×16 image resolution our model
achieves an accuracy of 82.16\% which is only 3.71\% lower than SKD, and for 32×32 image
resolution our model achieves an accuracy of 83.49\% which is only
6.42\% lower than SKD while in both cases, our model size is about
79× smaller and uses only 5400 pair of images as the training set. On the other hand,
SKD uses a pre-trained model on the VGGFace dataset \cite{parkhi2015deep} with 2.6 M images as its teacher model and train and fine-tunes its student model on the UMDFaces dataset \cite{bansal2017umdfaces} with 367,888
images. Based on these results, our model does have a competitive
performance for deployment in resource-constrained environments.

%%%%%%%%%%%%%%%%%%%%%%%%%%%%%%%%%%%%%%%%%%%%%
\begin{table*}
\caption{Low-Resolution Face verification results on LFW.}\label{table:lfw}
\begin{center}
\begin{tabular}{ccccc}
\hline
Model  & Resolution & \#Param. & training set (\#Img.)& Acc.(\%) \\ \hline
SKD without distillation &  16×16   & 0.79 M & UMDFaces (367,888)& 62.82 \\
SKD &      16×16   & 0.79 M & VGGFace (2.6 M), UMDFaces (367,888)& 85.87  \\
Ours &                      16×16   & 0.01 M & LFW (5400 pairs)& 82.16 \\
SKD without distillation&   32×32   & 0.79 M & UMDFaces (367,888)  & 70.23 \\
SKD &      32×32 & 0.79 M & VGGFace (2.6 M), UMDFaces (367,888)  & 89.72   \\
Ours &                      32×32 & 0.01 M & LFW(5400 pairs)& 83.30    \\ \hline
\end{tabular}
\end{center}
\end{table*}
%%%%%%%%%%%%%%%%%%%%%%%%%%%%%%%%%%%%%%%%%%%%%

\begin{figure}[t]
\begin{center}
\includegraphics[width=0.55\linewidth]{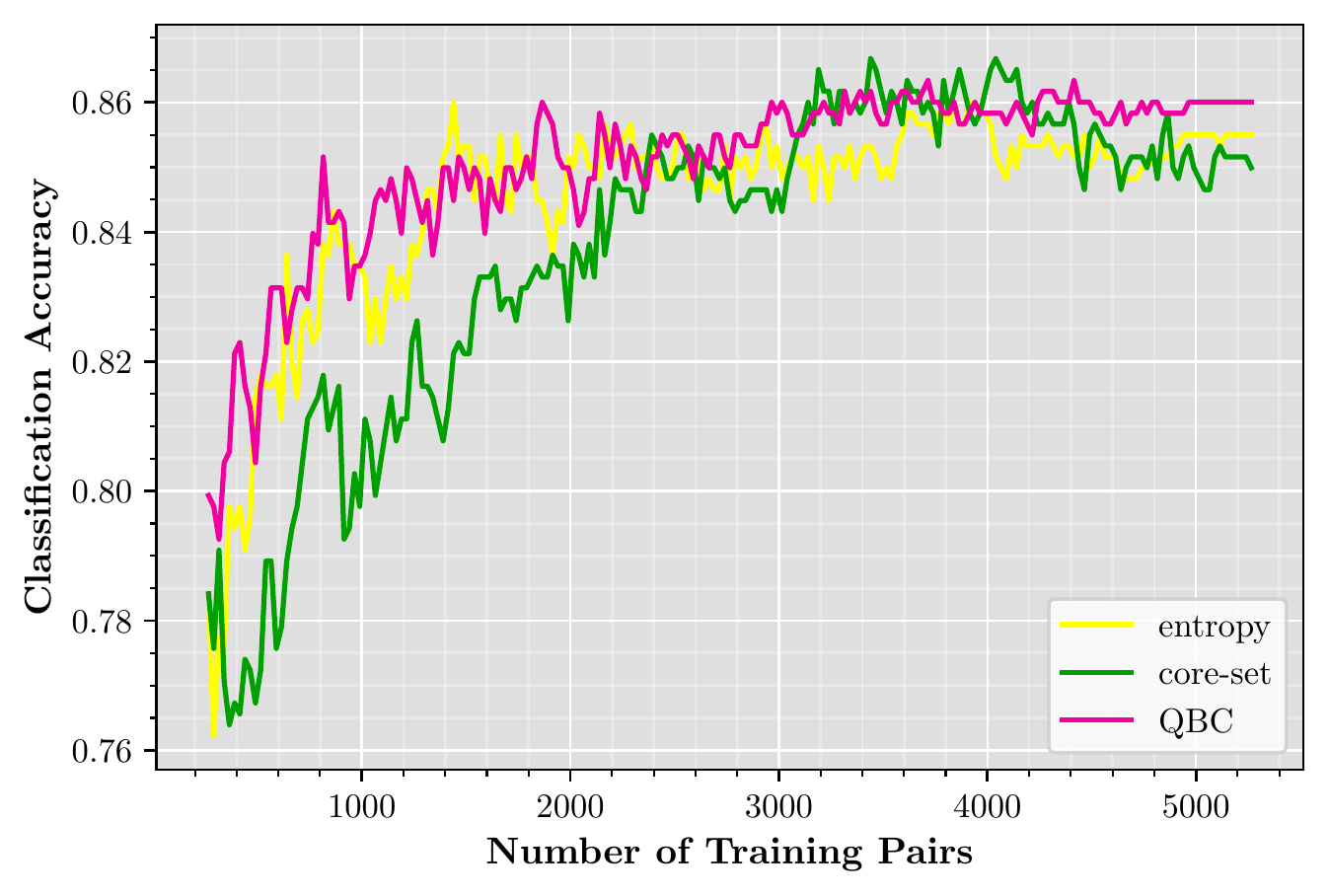}
\end{center}
\caption{Comparison of classification accuracy of three active learning
methods as a function of the number of training pairs.} 
\label{fig:activelearning}
\end{figure}

{\bf Experiment \#3.} In this experiment, we use the three active
learning methods which are introduced in Sec. \ref{sec:active} to
obtain the minimum required number of labeled training pairs without
significant loss in recognition accuracy.  We use the
first 90\% of the LFW training pairs as the initial pool of unlabeled
data ($D_u$) and the rest as the test data, then we randomly select 5\%
of the pool as $D_0$. The data augmentation step is removed.  For the
QBC algorithm, we use two different classifiers (LR and SVM) as the
committee of classifiers. We apply each active learning algorithm to
samples and compute the test accuracy versus the number of labeled
training pairs used for training the model. The result is shown in Fig.
\ref{fig:activelearning}. 

By comparing the performance of different algorithms, we see that QBC is
the most effective one for our model in the current experiment setting
as it has the fastest convergence in test accuracy and also the most
stable performance. For example, with this algorithm, our model can
achieve an accuracy of above 84\% using only 1465 training pairs.
Furthermore, our model achieves a stable accuracy of about 86\% with
only 3000 pairs. 

%%%%%%%%%%%%%%%%%%%%%%%%%%%%%%%%%%%%%%%%%%
%%%%%%%%%%%%%%%%%%%%%%%%%%%%%%%%%%%%%%%%%%

\subsection{Face Identification on CMU Multi-PIE}

We evaluate the effectiveness of our model for the face identification task
by conducting experiment on the Multi-PIE dataset. To the best of our
knowledge, no low-resolution face recognition model has reported results
on this dataset. Thus, we compare our model with high-resolution face
recognition models that have used this dataset for performance benchmarking. We use a setting mentioned in \cite{yim2015rotating} as Setting-1.
According to this setting, the first 150 identities are used for
training and the rest of the identities (151-250) from session 01 with
neutral expression are used for test so that there is no overlap between
training and test identities. As the gallery image, one frontal face
image with frontal illumination is used and the rest of the images in
the test set are selected as probes.  For model training, we generate
17,888 training pairs by pairing probe images with gallery images and
21,000 training pairs by randomly pairing images with each other.
Overall, 38,888 training pairs are used to train our model. 

Since our model is not proposed for handling pose variation, we report
results only for frontal and slightly non-frontal images
($\pm15^{\circ}$) in the dataset. The rank-1 face identification
accuracy is shown in Table. \ref{table:mpie}. Our model has competitive
and comparable performance although it uses low-resolution face images
and a small training set and has a small model size. For example, Light
CNN-29 has 12.637M parameters and uses 128×128 images while our model
size is 1,149× smaller and uses 32×32 images. It is worthwhile to comment
that TP-GAN in Table. \ref{table:mpie} also uses Light CNN-29 for feature
extraction from images so its model is larger than Light CNN-29. 

%%%%%%%%%%%%%%%%%%%%%%%%%%%%%%%%%%%%%%%%%%%%%
\begin{table}
\caption{Rank-1 identification rate (\%) for frontal and slightly
non-frontal face images ($\pm15^{\circ}$) in Setting-1.} 
\begin{center}
\begin{tabular}{ccc}
\hline
Method & Resolution & Acc.(\%)  \\ \hline
CPF \cite{yim2015rotating}& 60×60 & 89.45 \\ 
HPN \cite{ding2017pose} & 256×220 &84.23 \\ 
c-CNN Forest \cite{xiong2015conditional} & - &96.97 \\
Light CNN-29 \cite{Wu_2018} & 128×128 &99.78 \\
TP-GAN \cite{huang2017beyond} & 128×128 & 99.78 \\ 
Ours & 32×32 & 89.48 \\\hline
\end{tabular}
\end{center}
\label{table:mpie}
\end{table}
%%%%%%%%%%%%%%%%%%%%%%%%%%%%%%%%%%%%%%%%%%%%%

\subsection{Model size computation}

We compute the size of our model based on the information in Table
\ref{table:acc1}. Each of the two PixelHop++ systems in our model has three levels
and the c/w Saab transform is applied on 5x5 blocks with a stride of one in
each level. Since $E_C=E_F$, there are no leaf nodes in the first and
second levels. The first level of the $M_Y$ system has 18 intermediate nodes
and 7 discarded nodes (25-18). Thus, it has 25×18 variables for storing
PCA components and one bias parameter (451 parameters in
total). The second level has 119 intermediate nodes and 25×18-119
discarded nodes. For each of the 18 output channels of level one, the
Saab transform is applied separately. Initially, 25×18 nodes are
generated, and 119 of them are then selected as intermediate nodes based
on the energy threshold.  For each Saab transform, the DC kernel is known
and constant so we have 18 repetitive kernels in
the second level. As a result, we have 25×(119-18) parameters for
storing PCA components and 18 bias parameters which is in total 2,543
parameters. By the same token, there are 25×(233-119)+199 parameters in
the third level of $M_Y$. In the Pairwise Feature Generator - Y unit,
the mean and the standard deviation of each PixelHop++ output node is
stored. Hence, the size of the unit is 2×(18+119+233) equal to 740. The LR
classifier's size equals its input feature size plus one which
is 341 for $C_Y$. Likewise, the size of all subsystems is
computed and summarized in Table. \ref{table:sizecomp}. 

%%%%%%%%%%%%%%%%%%%%%%%%%%%%%%%%%%%%%%%%%%%%%
\begin{table}
\caption{The number of the parameters for different components of our model}\label{table:sizecomp}
\begin{center}
\begin{tabular}{cc}
\hline
subsystem & Num. of Param. \\ \hline
First hop - $M_Y$ & 451 \\ 
Second hop - $M_Y$ & 2,543 \\ 
Third hop - $M_Y$ & 2,969 \\
Pairwise Feat. Gen. - Y & 740 \\
LR Classifier - $C_Y$ & 341 \\
First hop - $M_{CrCb}$ & 476 \\ 
Second hop - $M_{CrCb}$ & 1,369 \\ 
Third hop - $M_{CrCb}$ & 1,348 \\ 
Pairwise Feat. Gen. - CrCb & 432 \\
LR Classifier - $C_{CrCb}$ & 242\\
Meta Classifier & 3 \\
Total & 10,914 \\\hline
\end{tabular}
\end{center}
\end{table}
%%%%%%%%%%%%%%%%%%%%%%%%%%%%%%%%%%%%%%%%%%%%%

\section{Conclusion and Future Work}\label{sec:conclusion}

A non-parametric solution targeting low-resolution face recognition for resource-constrained environments
was proposed in this work. Face biometric attributes (e.g., gender, ethnicity, age, etc.) prediction could be explored using a similar methodology. Furthermore,
the advantages of our solution may not be limited to resource-constrained
environments.  The same SSL principle could be generalized to
resource-rich environments and high-resolution images.  They are all interesting topics for future
research. 

\section{Acknowledgements}
This work was sponsored by US Army Research Office (ARO) under contract number W911NF1820218 through Carnegie Mellon University under subaward number 1130234-421349.

\bibliographystyle{unsrt}  
\bibliography{paper}  %%% Remove comment to use the external .bib file (using bibtex).


\end{document}